\documentclass{article}

\usepackage[preprint]{styles/neurips_2024}
\usepackage{graphicx}
\usepackage{subcaption}
\usepackage{wrapfig}
\usepackage{listings}
\usepackage{makecell}
\usepackage[utf8]{inputenc} 
\usepackage[T1]{fontenc}    
\usepackage{hyperref}       
\usepackage{url}            
\usepackage{booktabs}       
\usepackage{amsfonts}       
\usepackage{nicefrac}       
\usepackage{microtype}      
\usepackage{xcolor}         
\usepackage{amsmath}

\newcommand\extrafootertext[1]{%
    \bgroup
    \renewcommand\thefootnote{\fnsymbol{footnote}}%
    \renewcommand\thempfootnote{\fnsymbol{mpfootnote}}%
    \footnotetext[0]{#1}%
    \egroup
}

\title{Does Representation Matter? Exploring Intermediate Layers in Large Language Models}

\author{Oscar Skean$^{1*}$ \quad Md Rifat Arefin$^2$
\quad Yann LeCun$^{3, 4}$
\quad Ravid Shwartz-Ziv$^{3, 5}$\\
  $^1$University of Kentucky  \quad $^2$Mila \quad   $^3$New York University \quad  
  $^4$Meta FAIR \quad  
  $^5$Wand.AI
}

\begin{document}
\maketitle

\begin{abstract}

Understanding what defines a ``good'' representation in large language models (LLMs) is fundamental to both theoretical understanding and practical applications. In this paper, we investigate the quality of intermediate representations in various LLM architectures, including Transformers and State Space Models (SSMs). We find that intermediate layers often yield more informative representations for downstream tasks than the final layers. To measure the representation quality, we adapt and apply a suite of metrics—such as prompt entropy, curvature, and augmentation-invariance—originally proposed in other contexts. Our empirical study reveals significant architectural differences, how representations evolve throughout training, and how factors like input randomness and prompt length affect each layer. Notably, we observe a bimodal pattern in the entropy of some intermediate layers and consider potential explanations tied to training data. Overall, our results illuminate the internal mechanics of LLMs and guide strategies for architectural optimization and training.


\end{abstract}

\section{Introduction}
\label{sec:intro}

\extrafootertext{  *  correspondence to oscar.skean@uky.edu}

Large Language Models (LLMs) have revolutionized natural language processing by achieving remarkable performance across a wide range of tasks~\citep{muennighoff2022mteb, hendrycks2020mmlu}. Despite their success, understanding what constitutes a ``good'' representation within these models remains an open question. Specifically, how do representations at different layers contribute to downstream task performance, and how can we quantify their quality?

However, most previous studies have focused primarily on final-layer representations, often overlooking the potential of intermediate layers. Recent work suggests that intermediate layers may offer richer or more generalizable features for certain tasks~\citep{bordes2022guillotine, gurnee2023language, fan2024notalllayers}. These observations prompt a deeper investigation into the layer-wise behavior of LLMs.

In this paper, we explore the quality of representations across different layers of LLMs in various settings, including different model architectures (Transformers~\citep{vaswani2017attention} vs.\ State Space Models (SSMs)~\citep{mamba}), training checkpoints, input randomness, and prompt length. Our main contributions are:

\begin{itemize}
    \item We demonstrate that intermediate layers consistently provide better representations for downstream tasks than the final layers.
    \item We apply and adapt existing metrics—such as prompt entropy, curvature, and augmentation-invariance metrics—to quantify representation quality in LLMs.
    \item We analyze how these metrics vary across different settings, including architectural differences, training progression, input randomness, and prompt length.
\end{itemize}

Furthermore, we uncover significant differences in the behavior of these metrics between Transformers and SSMs. Notably, we observe a bimodal distribution in entropy within intermediate layers and investigate potential causes, such as the influence of training data examples.

Ultimately, our findings provide a deeper understanding of how internal representations develop in LLMs and offer practical guidance for model optimization. By illuminating the intricacies of intermediate layers, we pave the way for improved architectures, better training strategies, and more efficient utilization of LLM representations.

\begin{table}[!t]
\centering
\caption{MTEB Downstream Task Performance Using Representations from Different Layers}
\label{tab:downstream_performance}
\scalebox{0.6}{
\begin{tabular}{lccc}
\toprule
\textbf{Model} & \textbf{\makecell{Number of Tasks where Best Performance \\ is not in Last Layer}} & \textbf{Avg. Last Layer Performance} &  \textbf{Avg. Best Layer Performance}\\
\midrule
LLM2Vec 8B (Transformer) & 100\% & 64.7\% & 66.8\%\\
Pythia 410M (Transformer) & 96.6\% & 49.8\% & 53.3\% \\
Mamba 130M (SSM) & 100\% & 46.9\% & 50.9\% \\
\bottomrule
\end{tabular}
}
\end{table}

\section{Related Work}
\label{sec:related}

Understanding representations in neural networks has been a topic of extensive research. \citet{alain2016understanding} analyzed hidden representations to interpret neural networks' learning processes. \citet{raghu2017svcca} introduced Singular Vector Canonical Correlation Analysis (SVCCA) to compare representations across layers and networks, providing insights into learning dynamics. In the context of Transformers, \citet{liu2019linguistic} studied the linguistic knowledge captured at different layers, finding that lower layers encode more syntactic information while higher layers capture semantic features. Similarly, \citet{jin2024conceptdepth} showed that semantic concepts are learned in intermediate layers and proposed a layer-wise probing technique to identify the specific layers where these concepts are formed. On the other hand, state-space models have been less explored in this regard. \citet{mamba} introduced Mamba, an SSM architecture capable of handling long sequences efficiently. However, comparative studies between SSMs and Transformers at the representation level remain scarce.

Metrics like entropy and curvature have been used in other contexts to analyze representations. \citet{shwartz2017opening, shwartz2022information} discussed the Information Bottleneck principle, suggesting that networks learn to compress representations. \citet{hosseini2024curvature} introduced curvature as a measure of representational dynamics in recurrent networks. Several works in the vision domain have proposed unsupervised representation quality metrics that are strongly correlated with accuracy on downstream tasks~\citep{garrido2023rankme, agrawal2022alphareq, thilak2023lidar}. Notably, the RankMe measure from \citet{garrido2023rankme} can be shown to be a measure of entropy known as matrix-based entropy, which we use in our analysis.

Our work bridges these areas by applying and adapting such metrics to LLMs, providing a novel perspective on representation quality across architectures and training stages.

\section{Methodology}
\label{sec:methodology}

\subsection{Notation}
We consider a batch of $N$ samples, each represented by a $D$-dimensional vector. Let $\mathbf{Z} \in \mathbb{R}^{N \times D}$ be the matrix of representations, where $z_i$ denotes the $i$-th row of $\mathbf{Z}$. For a matrix $\mathbf{M}$, we use $\lambda_i(\mathbf{M})$ to denote its $i$-th largest eigenvalue, and $\operatorname{tr}(\mathbf{M})$ to denote its trace. When dealing with sequences, we let $\mathbf{x} \in \mathbb{R}^{L \times d}$ represent the input sequence and $\mathbf{y} \in \mathbb{R}^{L \times d}$ the output sequence, where $L$ is the sequence length and $d$ is the feature dimension.


\subsection{Architectures}
\label{sec:architectures}

We compare two main types of architectures: Transformer-based models \citep{vaswani2017attention} and State Space Models (SSMs) \citep{mamba}.

\textbf{Transformers:} Transformers use self-attention layers to capture long-range dependencies within the input. By computing attention weights between tokens, they can integrate global context at every layer and scale effectively to large inputs.

\textbf{State Space Models (SSMs):} SSMs represent sequence processing using linear state transitions combined with gating mechanisms. They offer efficient handling of long sequences with linear time and memory complexity, making them a promising alternative to Transformers.

For further details on each architecture and their configurations, see Appendix~\ref{appendix:architectures}.




\subsection{Representation Evaluation Metrics}
\label{sec:metrics}

We use two categories of metrics to evaluate representation quality: token embedding diversity metrics and augmentation-invariance metrics.






\subsubsection{Token Embedding Diversity Metrics}
\label{sect:token-embedding-diversity-metrics}

Token embedding diversity metrics evaluate the variability and richness of the representations at the token level within a single sequence. These metrics are designed to capture how distinctively each token is represented within the context of the entire prompt, providing insight into how effectively the model encodes information and differentiates between different parts of the input.

\paragraph{Prompt Entropy:}

Following \citet{wei2024large}, we use the $\alpha$-order matrix-based entropy \citep{giraldo2014measures} as a surrogate for Rényi entropy. For a sequence of token representations $\mathbf{Z} \in \mathbb{R}^{L \times d}$, the Gram matrix is $\mathbf{K_Z} = \mathbf{Z}\mathbf{Z}^\top$. The entropy is:

\begin{equation}
\label{eq:matrix-based-entropy}
    S_{\alpha}(\mathbf{Z}) = \frac{1}{1 - \alpha} \log \left( \sum_{i=1}^{L} \left( \frac{\lambda_i(\mathbf{K_Z})}{\operatorname{tr}(\mathbf{K_Z})} \right)^{\alpha} \right).
\end{equation}

In this context, prompt entropy quantifies the degree of diversity and dispersion in token embeddings within a single sequence. Higher entropy values indicate that the model preserves more nuanced and varied token-level information. Conversely, lower entropy suggests that the model compresses the input representations into fewer dimensions or patterns. As such, prompt entropy provides a useful measure of how well the model maintains complexity and richness in its intermediate representations.

Unless otherwise specified, we use the limit case $\alpha=1$ in our calculations. At this limit, the metric is equivalent to the RankMe measure defined in \cite{garrido2023rankme}. We explore the effects of different $\alpha$ values in Appendix \ref{appendix:entropy}. For a more in-depth examination of prompt entropy, refer to Appendix~\ref{sect:appendix-prompt-entropy}.

\paragraph{Curvature}

As introduced by \citet{hosseini2024curvature}, curvature measures how rapidly the direction between two adjacent token embedding vectors changes. Define their difference as $\mathbf{v}_k = \mathbf{z}_{k+1} - \mathbf{z}_k$. The average curvature of a prompt is:

\begin{equation}
    \bar{C} = \frac{1}{L-2} \sum_{k=1}^{L-2} \arccos\left( \frac{\mathbf{v}_{k+1}^\top \mathbf{v}_k}{\|\mathbf{v}_{k+1}\| \|\mathbf{v}_k\|} \right).
\end{equation}

\subsubsection{Augmentation Invariance Metrics}

These metrics measure how consistently a model represents a prompt when it is perturbed or augmented. Because augmentations may change prompt length, we average all token embeddings to form a single vector per prompt.

 Because augmentation may change the prompt length, the token embedding diversity metrics described in~\ref{sect:token-embedding-diversity-metrics} are no longer suitable. Instead, we average all token embeddings to form a single vector per prompt and use the metrics described below to measure the similarity between two augmentations of the same prompt.

 Let $Z_1 \in \mathbb{R}^{N \times D}$ and $Z_2 \in \mathbb{R}^{N \times D}$ represent two augmented sets of $N$ prompts, where the $i$-th row in both corresponds to the same original prompt. Details on the augmentation process are in Appendix~\ref{appendix:prompt-augmentation}.


\paragraph{InfoNCE}

InfoNCE~\citep{oord2018representation} provides a mutual information lower bound between paired augmentations. Lower InfoNCE loss suggests that augmentations of the same prompt map to similar representations, indicating invariance to perturbations. This loss is widely used to train augmentation-invariant networks in self-supervised learning for vision and is well-suited to capturing the semantic similarity underlying the augmented prompts~\citep{chen2020simclr, chen2020mocov2, shwartz2024compress, NEURIPS2023_b63ad8c2}.

\paragraph{DiME}

DiME~\citep{skean2023dime} compares the alignment of paired samples to that of randomly paired samples. Similar to InfoNCE, it is used to estimate the mutual information between two augmented sets of prompts. DiME is grounded in the matrix-based entropy defined in Eq.~\ref{eq:matrix-based-entropy}. In essence, it quantifies how closely the pairings in $(Z_1, Z_2)$ resemble each other, compared to pairings of $(Z_1, \Pi Z_2)$ for a permutation matrix $\Pi$. Higher DiME values imply that correct augmentation pairs yield representations that are significantly more similar than random pairings, indicating stronger augmentation invariance.

\paragraph{LiDAR}
LiDAR~\citep{thilak2023lidar} employs a linear discriminant analysis (LDA) framework to assess how well augmentations of a single prompt cluster together. Each prompt is considered a separate class, with its augmentations serving as class samples. By examining the variances of the linear discriminant components, LiDAR quantifies the tightness of these clusters. Higher LiDAR scores indicate that augmentations belonging to the same prompt form more coherent groups, reflecting stronger invariance.

To compute the LDA matrix, LiDAR uses augmentations to construct the class scatter matrix. In our setup, we use $N$ classes (one for each prompt) and $J=16$ samples per class. This is a larger sample size than the $J=2$ used in DiME or InfoNCE, reflecting the more complex requirements of computing the LDA matrix.






\section{Experiments}
\label{sec:experiments}

\subsection{Intermediate Layers Provide Better Representations for Downstream Embedding Tasks}

We begin by evaluating representations at each model layer on a suite of downstream tasks from the Massive Text Embedding Benchmark (MTEB)~\citep{muennighoff2022mteb}. MTEB is designed to test the performance of LLMs on various embedded tasks. We chose 32 tasks covering classification, clustering, and
re-ranking. We use three models: Pythia 410M, Mamba 130M, and LLM2Vec-unsup-simcse~\citep{behnamghader2024llm2vec}.


Our findings indicate that intermediate layers consistently outperform the final layer across all three architectures (Table~\ref{tab:downstream_performance}). Selecting the best-performing intermediate layer yields at least a 2\% improvement in average accuracy compared to using the last layer. While prior work~\citep{fan2024notalllayers} noted similar trends for generation tasks, our results extend this observation to embedding-based evaluations.


\subsection{Downstream Performance and Entropy Are Negatively Correlated}

We next examine how prompt entropy relates to downstream performance on the Massive Multitask Language Understanding (MMLU) benchmark~\citep{hendrycks2020mmlu}, which tests comprehensive knowledge across 57 diverse subjects, covering topics from elementary mathematics to professional law.

We compare two similarly sized models, Llama3-8B and Mamba2-8B. Despite having the same parameter count, Llama3 achieves $63.85 \pm 0.38\%$ accuracy, far surpassing Mamba2’s $26.76 \pm 0.37\%$. We hypothesize that Llama3’s intermediate layers compress information more effectively, helping it discard irrelevant details and focus on task-relevant features. As shown in Figure~\ref{fig:metrics-across-architectures}, the correlation between intermediate-layer entropy and MMLU performance in Llama3 is strongly negative (-0.43 between the second and later layers) (Figure~\ref{fig:with_logit22}). In contrast, Mamba2 shows no such relationship, nor evidence of similar compression (Figure~\ref{fig:with_logit11}).


\subsection{Experimental Setup for Evaluating Representation Quality}
\label{sect:metrics-experiments}

We now apply the metrics from Section~\ref{sec:metrics} to quantify representation quality layer-by-layer. Our experiments span both Transformers, SSMs, and Pythia~\citep{biderman2023pythia}, including various scales. We utilize two datasets: WikiText-103~\citep{merity2016pointer}, representing general text, and an instruction-based medical dataset~\citep{ruslanmv2024} for more specialized content. This setup allows us to probe how architectural choices and input complexity affect internal representations.


\subsubsection{Architectural Differences}
\label{sect:metrics-across-architectures}

\begin{figure}[!t]
    \centering
    \begin{subfigure}[b]{0.28\textwidth}
        \centering
        \includegraphics[width=\textwidth]{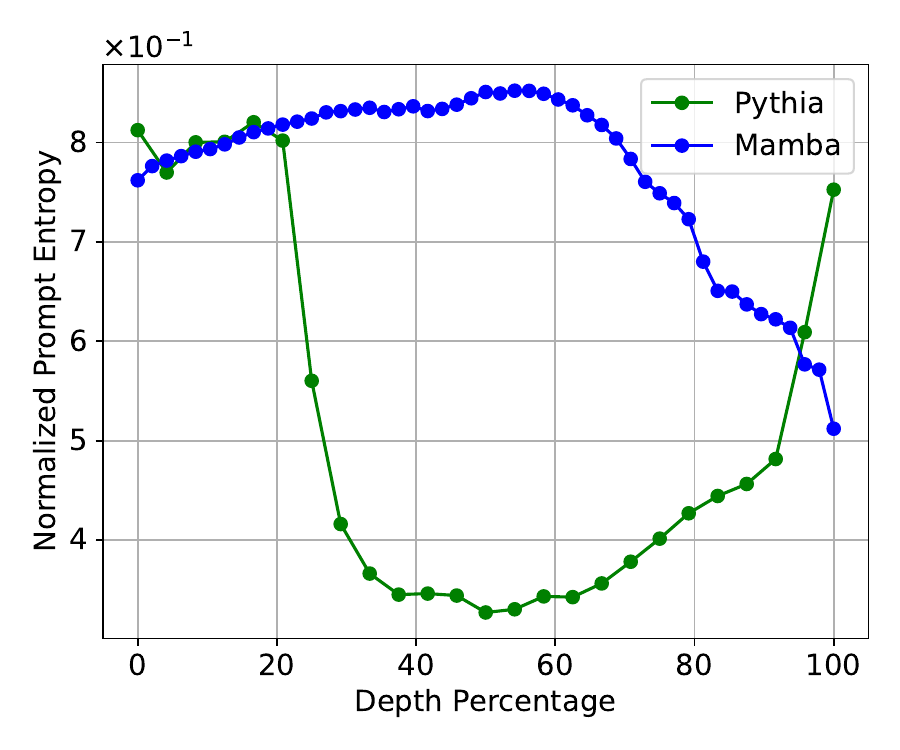}
        \caption{Prompt Entropy}
    \end{subfigure}%
    \hspace{0.04\textwidth}
    \begin{subfigure}[b]{0.28\textwidth}
        \centering
        \includegraphics[width=\textwidth]{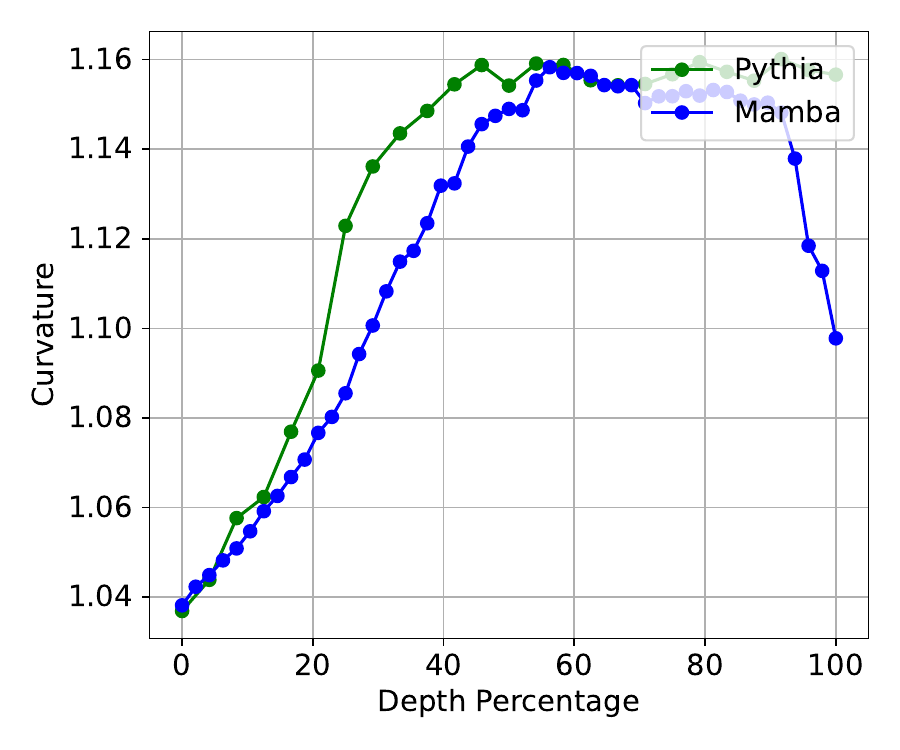}
        \caption{Curvature}
    \end{subfigure}%
    \hspace{0.04\textwidth}
    \begin{subfigure}[b]{0.28\textwidth}
        \centering
        \includegraphics[width=\textwidth]{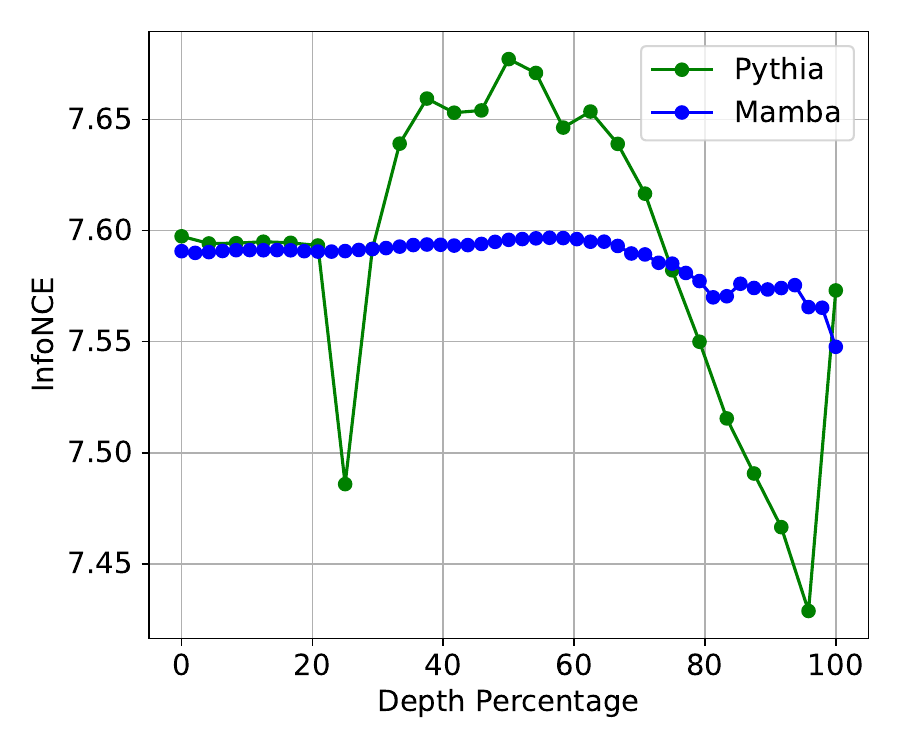}
        \caption{infoNCE}
    \end{subfigure}
    
    \vspace{0.5cm} 
    \begin{subfigure}[b]{0.28\textwidth}
        \centering
        \includegraphics[width=\textwidth]{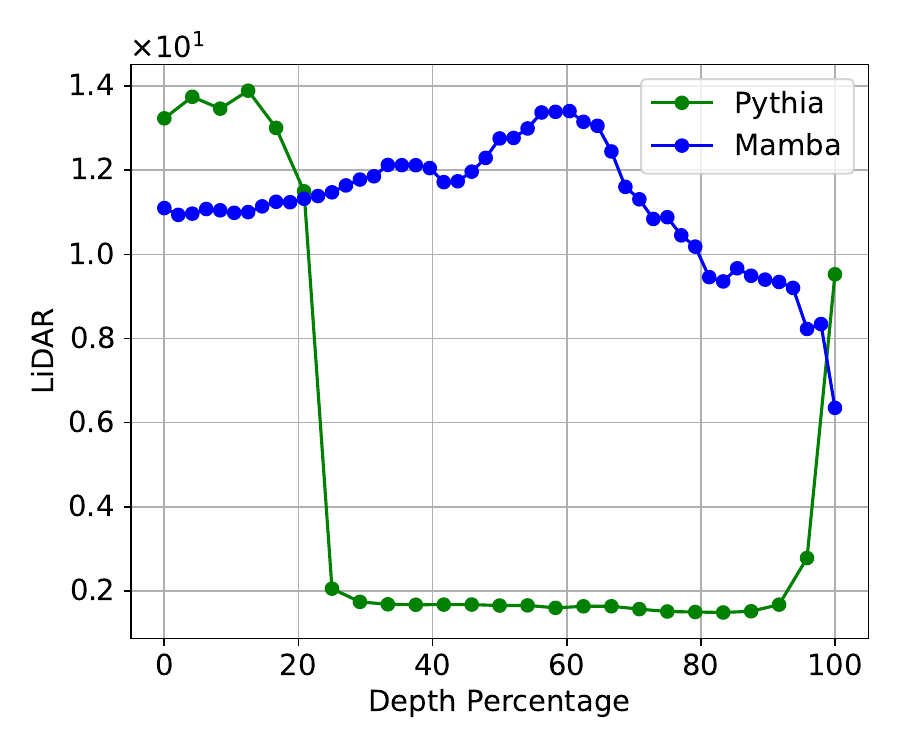}
        \caption{LiDAR}
    \end{subfigure}%
    \hspace{0.04\textwidth}
    \begin{subfigure}[b]{0.28\textwidth}
        \centering
        \includegraphics[width=\textwidth]{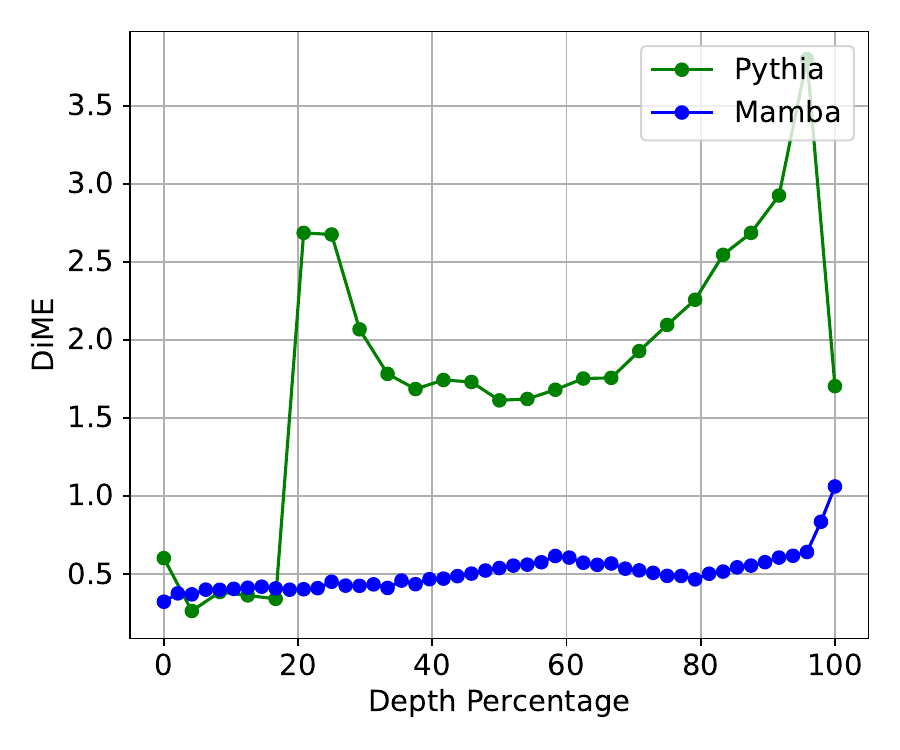}
        \caption{DiME}
    \end{subfigure}%
    \hspace{0.04\textwidth}
    \begin{subfigure}[b]{0.28\textwidth}
        \centering
        \includegraphics[width=\textwidth]{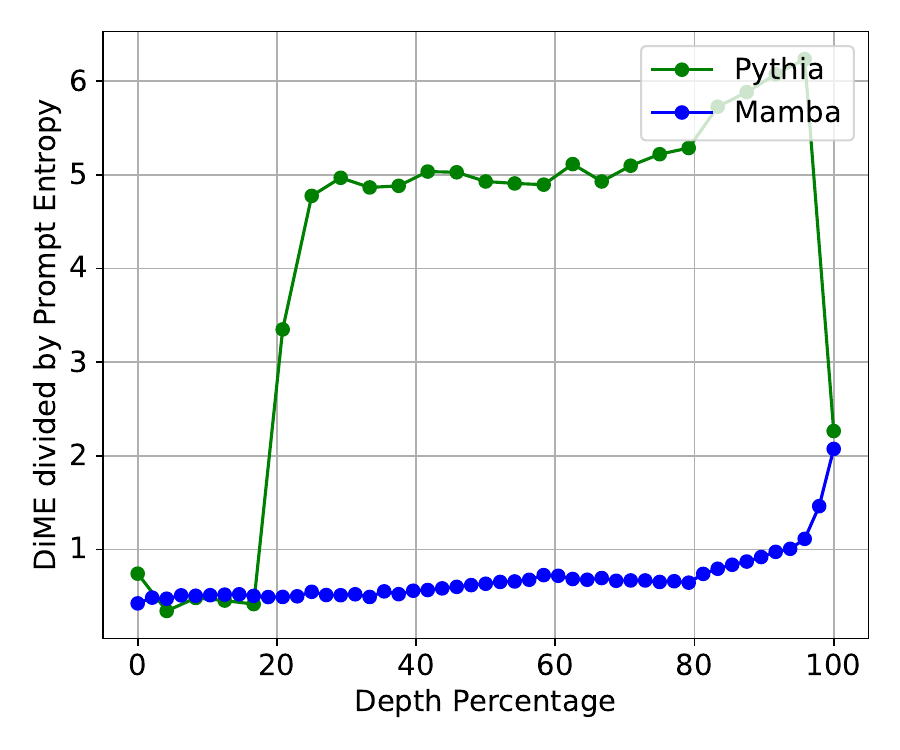}
        \caption{DiME divided by Pr. Ent.}
    \end{subfigure}
  \caption{\textbf{Pythia’s intermediate layers show pronounced changes in representation quality metrics, while Mamba’s remain more stable.} Representation evaluation metrics across layers in Pythia 410M and Mamba 370M architectures. The x-axis denotes model depth as a percentage, allowing fair comparison between models with different layer counts.}
  \label{fig:metrics-across-architectures}
\end{figure}

Our analysis reveals notable differences in representation quality between Transformer-based architectures (e.g., Pythia) and SSMs (e.g., Mamba). Figure \ref{fig:metrics-across-architectures} compares entropy, InfoNCE, LiDAR, and DiME metrics as a function of model depth, normalized to allow fair comparisons across models with different numbers of layers.


For entropy and LiDAR, Pythia shows a pronounced decrease at intermediate layers, suggesting greater information compression and consolidation. In contrast, Mamba maintains more stable values, indicating less compression in its intermediate representations. Meanwhile, Mamba exhibits lower DiME and InfoNCE values than Pythia, implying reduced variability in its intermediate-layer representations.


Overall, these metric shifts are more pronounced in Pythia than in Mamba, suggesting that Pythia undergoes stronger representational transformations at intermediate depths. By comparison, Mamba’s representations remain more uniform across layers. These differences may influence how each model encodes and leverages information for downstream tasks.

\subsubsection{Impact of Training Progression}

\begin{figure}[!t]
    \centering
    \begin{subfigure}[b]{0.28\textwidth}
        \centering
        \includegraphics[width=\textwidth]{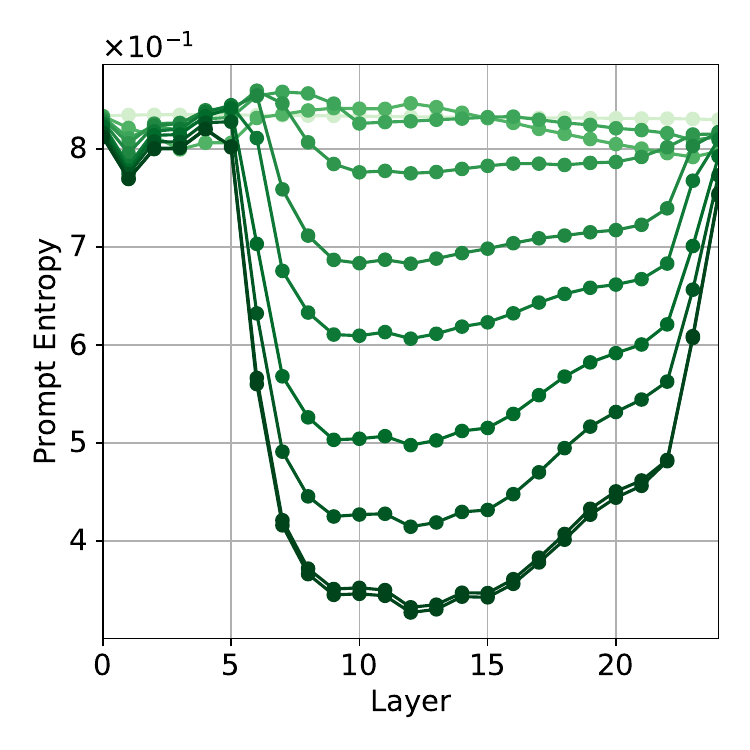}
        \caption{Prompt Entropy}
    \end{subfigure}%
    \hspace{0.04\textwidth}
    \begin{subfigure}[b]{0.28\textwidth}
        \centering
        \includegraphics[width=\textwidth]{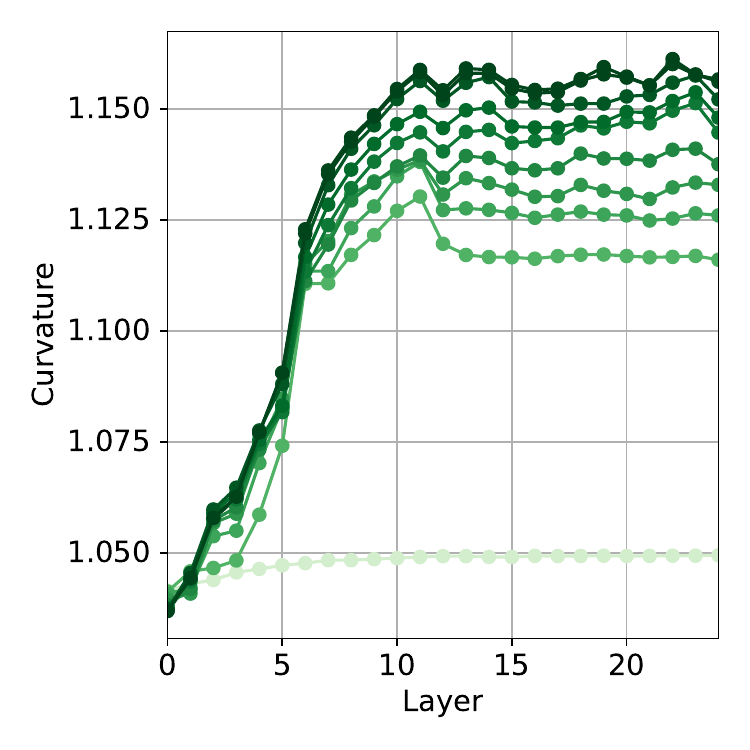}
        \caption{Curvature}
    \end{subfigure}%
    \hspace{0.04\textwidth}
    \begin{subfigure}[b]{0.35\textwidth}
        \centering
        \includegraphics[width=\textwidth]{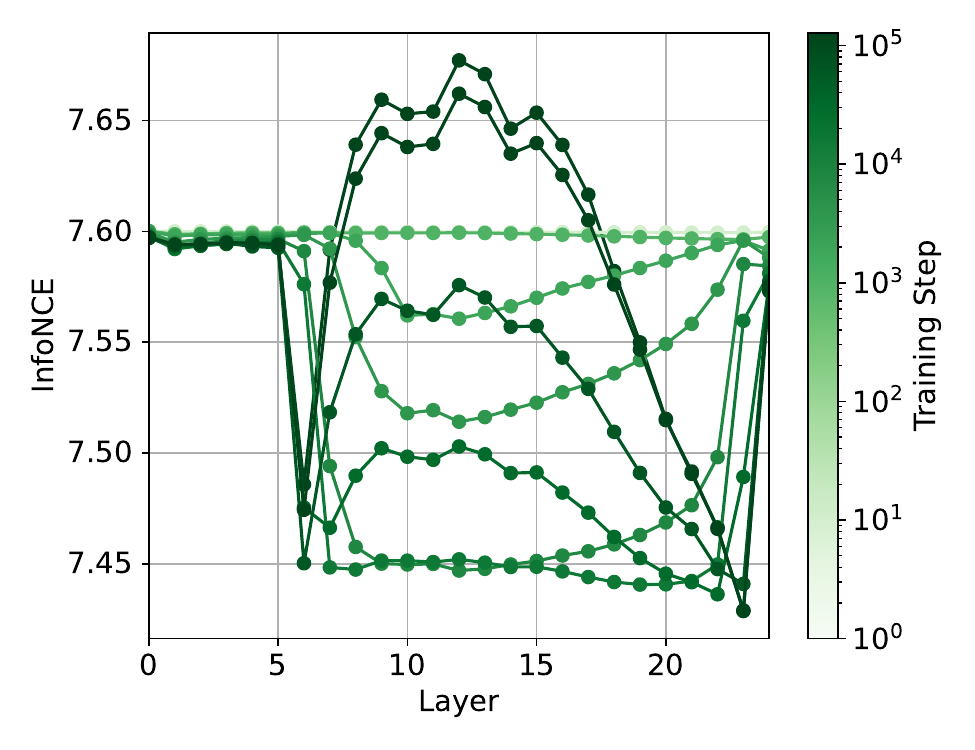}
        \caption{infoNCE}
    \end{subfigure}
    
    \vspace{0.5cm} 
    \begin{subfigure}[b]{0.28\textwidth}
        \centering
        \includegraphics[width=\textwidth]{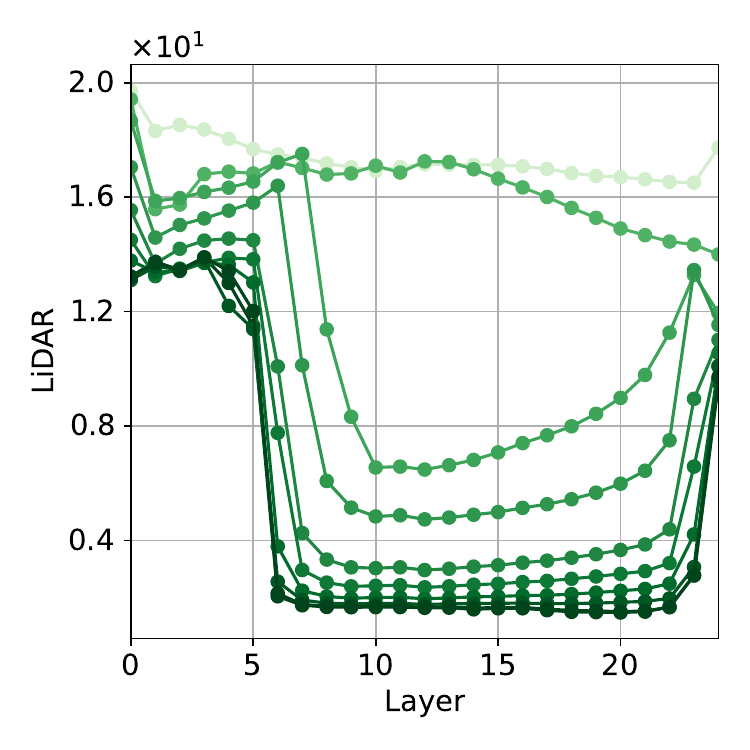}
        \caption{LiDAR}
    \end{subfigure}%
    \hspace{0.04\textwidth}
    \begin{subfigure}[b]{0.28\textwidth}
        \centering
        \includegraphics[width=\textwidth]{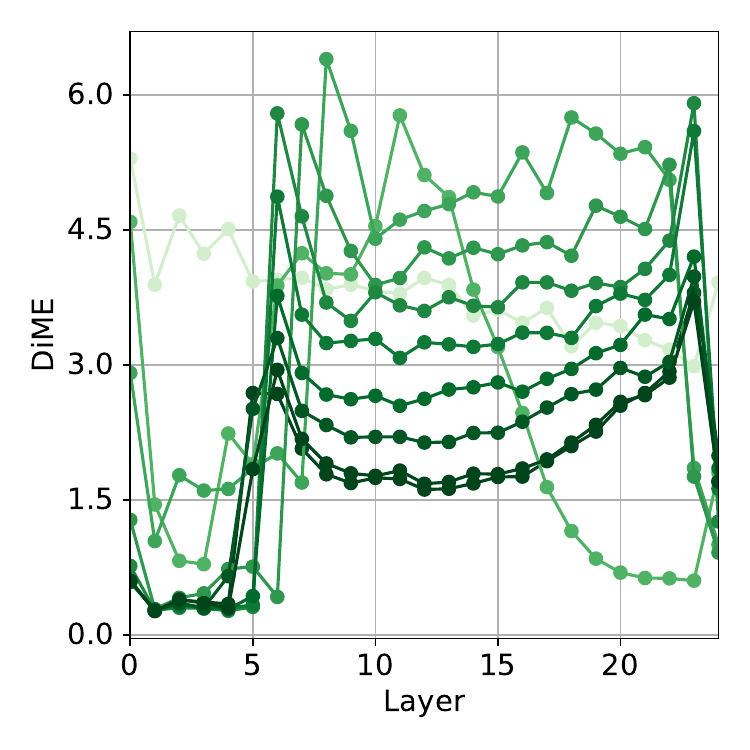}
        \caption{DiME}
    \end{subfigure}%
    \hspace{0.04\textwidth}
    \begin{subfigure}[b]{0.35\textwidth}
        \centering
        \includegraphics[width=0.8\textwidth]{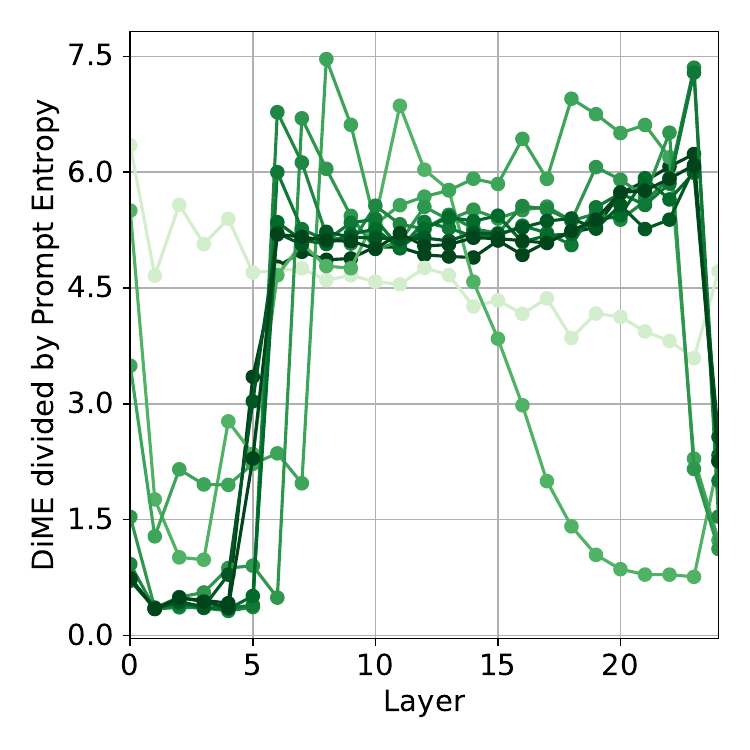}
        \caption{DiME divided by Pr. Ent.}
    \end{subfigure}
    
    \caption{\textbf{Training effects are most pronounced in the intermediate layers.} Representation metrics across layers at different training checkpoints (steps 1 to 143k). The x-axis is the depth percentage of the model, showing how training influences different layers, particularly those at intermediate depths.}
    \label{fig:metrics_across_training}
\end{figure}

To examine how representation quality evolves over the course of training, we analyze Pythia's representations at various checkpoints. Figure \ref{fig:metrics_across_training} reports several evaluation metrics across layers from the initial training step up to step 143k, sampled on a logarithmic scale.


The results show that the most significant changes occur in the intermediate layers. As training progresses, prompt entropy in these layers decreases, indicating that the model is learning to compress and abstract input information more efficiently. In contrast, the InfoNCE metric peaks in the intermediate layers, suggesting that the representations become more distinct. Meanwhile, LiDAR and DiME values both decline, reflecting a reduction in variability along certain representational dimensions.


Interestingly, the metrics for the earliest layers remain relatively stable throughout training. This observation aligns with the detokenization hypothesis proposed by \citep{lad2024remarkable}, which posits that initial layers primarily handle the mapping of raw input tokens into an initial embedding space. Their roles appear to solidify early on, exhibiting less ongoing change than the intermediate layers.



\subsubsection{Prompt Entropy under Extreme Input Conditions}

To gain deeper insights into how prompt entropy behaves under various input perturbations, we investigate the impact of extreme prompt modifications on the model's internal representations. Specifically, we analyze how prompt entropy evolves across different layers of the Pythia 410M model when subjected to high levels of token repetition, randomness, or increased prompt length.


We design three types of extreme prompts:

\begin{enumerate}

    \item \textbf{Prompts with Increasing Token Repetition}: We select 1,000 standard prompts from the WikiText dataset and randomly replace tokens with a fixed token from the prompt at varying probabilities $p$. As $p$ increases, the amount of repetition in the prompt increases.

    \item \textbf{Prompts with Increasing Token Randomness}: We introduce randomness by randomly substituting tokens in the prompts with arbitrary tokens from the vocabulary at varying probabilities $p$. Higher values of $p$ correspond to greater randomness in the prompts.

    \item \textbf{Random Prompts of Increasing Length}: We generate random prompts by sampling tokens uniformly from the vocabulary, creating prompts of varying lengths $T$.
\end{enumerate}

\begin{figure}[!t]
    \centering
    \begin{subfigure}[b]{0.32\textwidth}
        \centering
        \includegraphics[width=\linewidth]{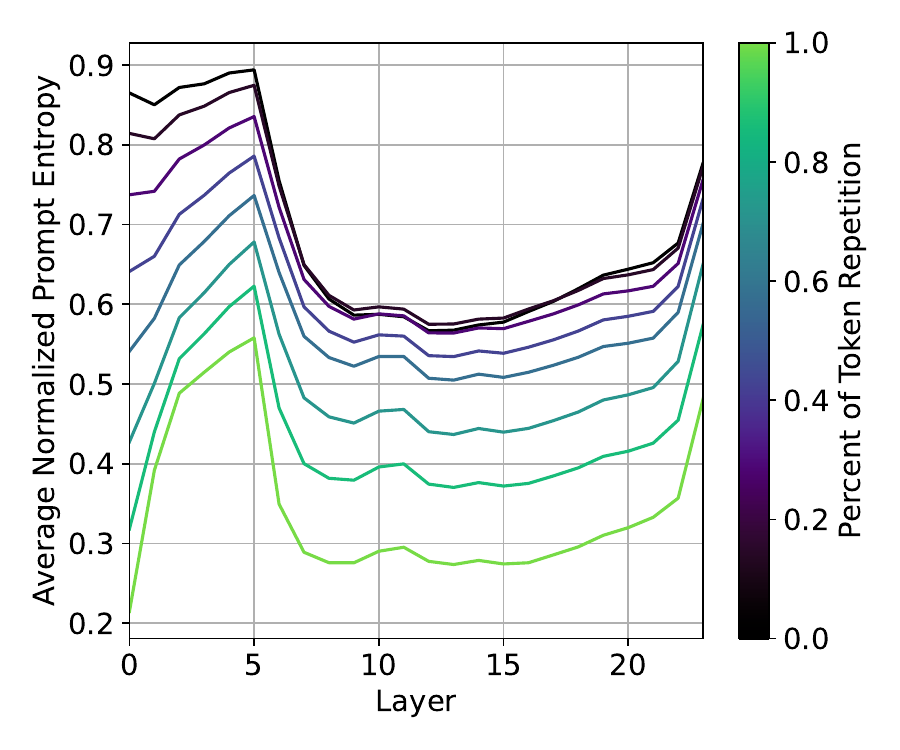}
        \caption{Repetition}
        \label{fig:pythia_increasing_repetition}
    \end{subfigure}\hfill
    \begin{subfigure}[b]{0.32\textwidth}
        \centering
        \includegraphics[width=\linewidth]{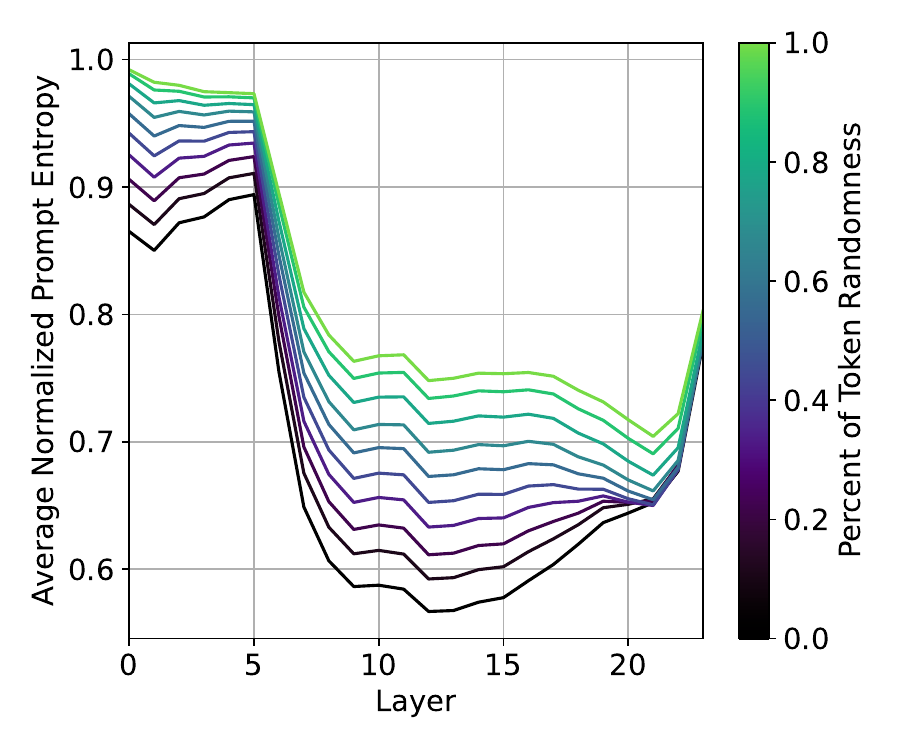}
        \caption{Randomness}
        \label{fig:pythia_increasing_randomness}
    \end{subfigure}\hfill
    \begin{subfigure}[b]{0.32\textwidth}
        \centering
        \includegraphics[width=\linewidth]{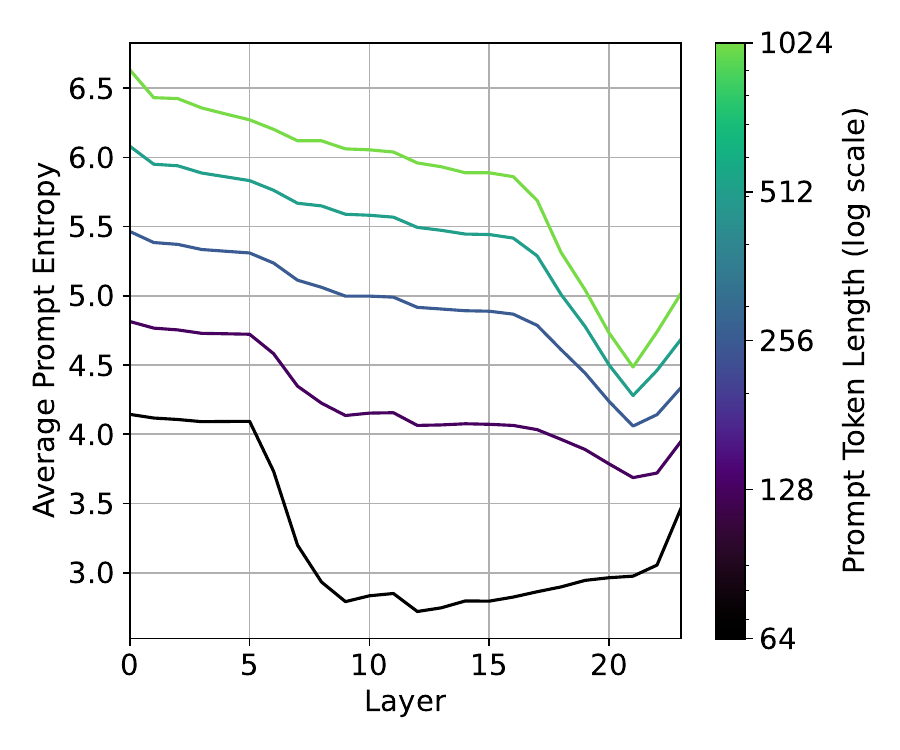}
        \caption{Random Prompt Length}
        \label{fig:prompt-random-raw}
    \end{subfigure}
    \caption{\textbf{Prompt entropy across layers of Pythia 410M under various extreme input conditions.} (a) Increasing token repetition leads to decreased entropy in intermediate layers. (b) Increasing token randomness results in higher entropy, especially in initial layers. (c) Unnormalized prompt entropy increases with prompt length due to the larger number of tokens. These results demonstrate how the model's internal representations adapt to different types of input perturbations.}
    \label{fig:pythia-increasing-intensity}   
\end{figure}

Figure \ref{fig:pythia-increasing-intensity} displays both normalized and unnormalized prompt entropy across different layers for each type of extreme prompt. The key observations from this analysis are:


\textbf{1. Increasing token repetition reduces entropy in intermediate layers.} As the probability $p$ of token repetition rises, the model compresses redundant information, leading to lower entropy values in the middle layers. This compression indicates that the model effectively recognizes and encodes repetitive patterns within the input.

\textbf{2. Increasing token randomness elevates entropy, particularly in initial layers.} Introducing random tokens enhances the diversity of token representations, resulting in higher entropy values. The initial layers exhibit the most significant increases, suggesting that these layers are more sensitive to input noise and variability.


\textbf{3. Prompt length influences entropy in Both normalized and unnormalized Forms.} Unnormalized entropy naturally grows with prompt length due to the increased number of tokens. Although not displayed, normalized entropy demonstrates sublinear growth, implying that each additional token contributes progressively less to the overall diversity as the prompt lengthens.


These findings illustrate that extreme input conditions distinctly affect the model's internal representations, especially within intermediate layers. The varying compression and encoding behaviors based on the nature of input perturbations provide valuable insights into the model's processing mechanisms and its capacity to maintain or reduce information complexity under different scenarios.


\subsection{Bimodal Behavior in Prompt Entropy}

During our analysis of average prompt entropy across different layers, we identified an intriguing phenomenon: a distinct bimodal distribution of entropy values in certain layers of Transformer models, which was absent in SSMs. Figure \ref{fig:all-models-bimodal} presents the entropy distributions for both the WikiText and AI-Medical-Chatbot datasets~\citep{ruslanmv2024}. Notably, the AI-Medical-Chatbot dataset exhibits a pronounced bimodal distribution in the middle layers of Transformer models. This suggests that the model processes some prompts in fundamentally different ways at these intermediate stages. To investigate the underlying causes of this bimodality, we conducted several experiments detailed in Appendix \ref{appendix:bimodal-investigation}. Our findings indicate that factors such as prompt length, semantic complexity, and overlap with training data do not account for this behavior. Consequently, the root cause of the bimodal entropy distribution remains an open question.


\section{Discussion and Conclusion}
\label{sec:discussion_conclusion}

In this work, we explored the representation quality of intermediate layers in LLMs, providing insights into their critical role in downstream task performance. By applying a diverse set of evaluation metrics, including prompt entropy, curvature, InfoNCE, LiDAR, and DiME, we highlighted distinct behaviors in Transformer-based architectures and SSMs. Our findings demonstrate that intermediate layers often outperform final layers in representation quality, underscoring their significance for feature extraction and transfer learning.


Transformers exhibited greater representational variability and information compression within intermediate layers, whereas SSMs displayed more stable and consistent representations. This suggests differing strategies in encoding information, with Transformers excelling in adaptability and SSMs prioritizing robustness. Furthermore, the training analysis revealed that the most substantial improvements in representation quality occur in intermediate layers, reinforcing their importance in learning dynamics.


Our investigation into extreme input conditions revealed that intermediate layers play a pivotal role in adapting to diverse input scenarios, with distinct responses to token repetition, randomness, and prompt length. Additionally, the observation of bimodal entropy distributions in intermediate layers of Transformer models remains an open question, offering avenues for further research.



In conclusion, our research advances the understanding of internal representation dynamics in LLMs, highlighting the pivotal role of intermediate layers and the distinct behaviors of different architectures. These findings not only enhance the theoretical knowledge of model representations but also provide practical guidance for optimizing model design, training, and application. Future work should delve deeper into the causes of phenomena such as bimodal entropy distributions and explore the development of new metrics specifically tailored to LLMs to further enhance representation evaluation.


\bibliographystyle{styles/neurips_2024}
\bibliography{strings, references}

\newpage
 \appendix



\begin{figure}[!h]
        \centering
        \begin{subfigure}[b]{0.8\textwidth}
                \centering
                \includegraphics[width=\linewidth]{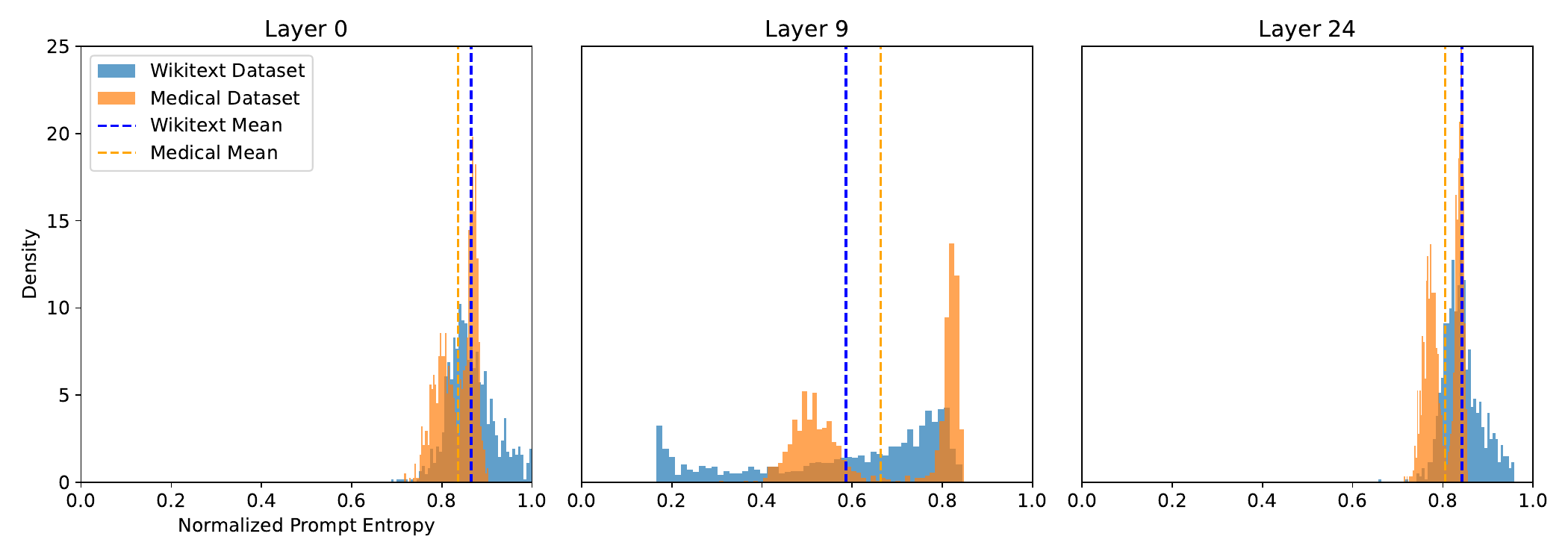}
                \caption{Pythia 410M}
        \end{subfigure}
        \begin{subfigure}[b]{0.8\textwidth}
              \centering
              \includegraphics[width=\linewidth]{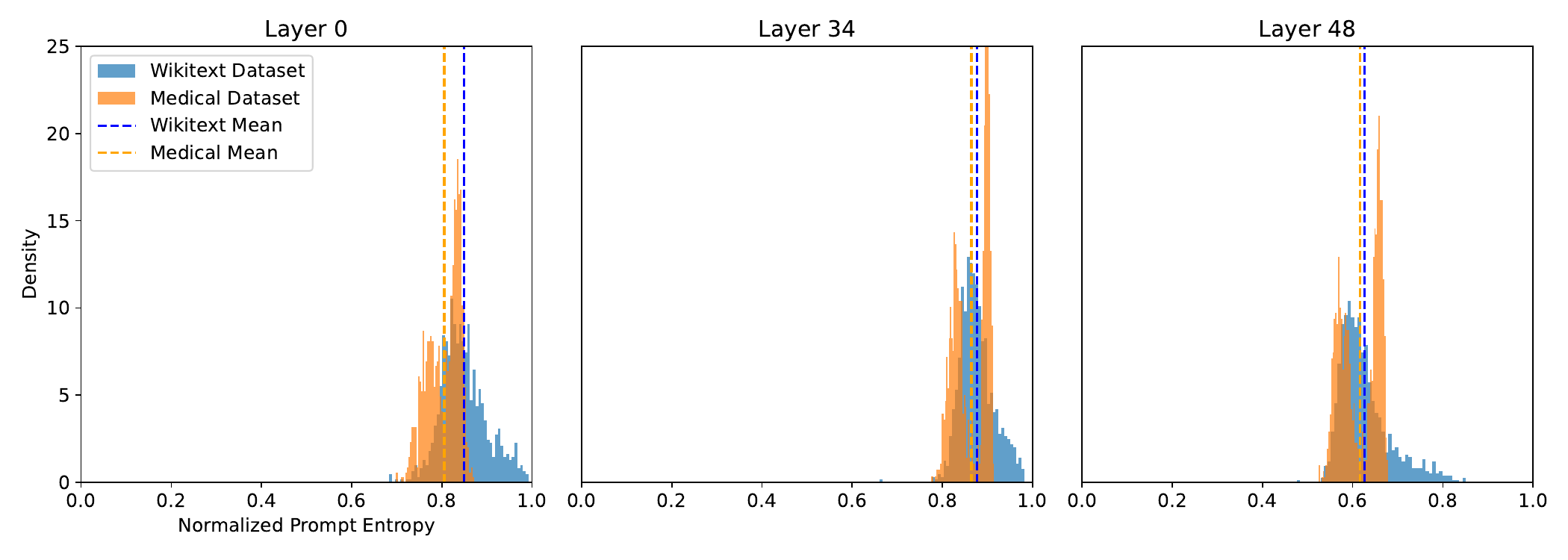}
                \caption{Mamba 370M}
        \end{subfigure}
        \begin{subfigure}[b]{0.8\textwidth}
              \centering
              \includegraphics[width=\linewidth]{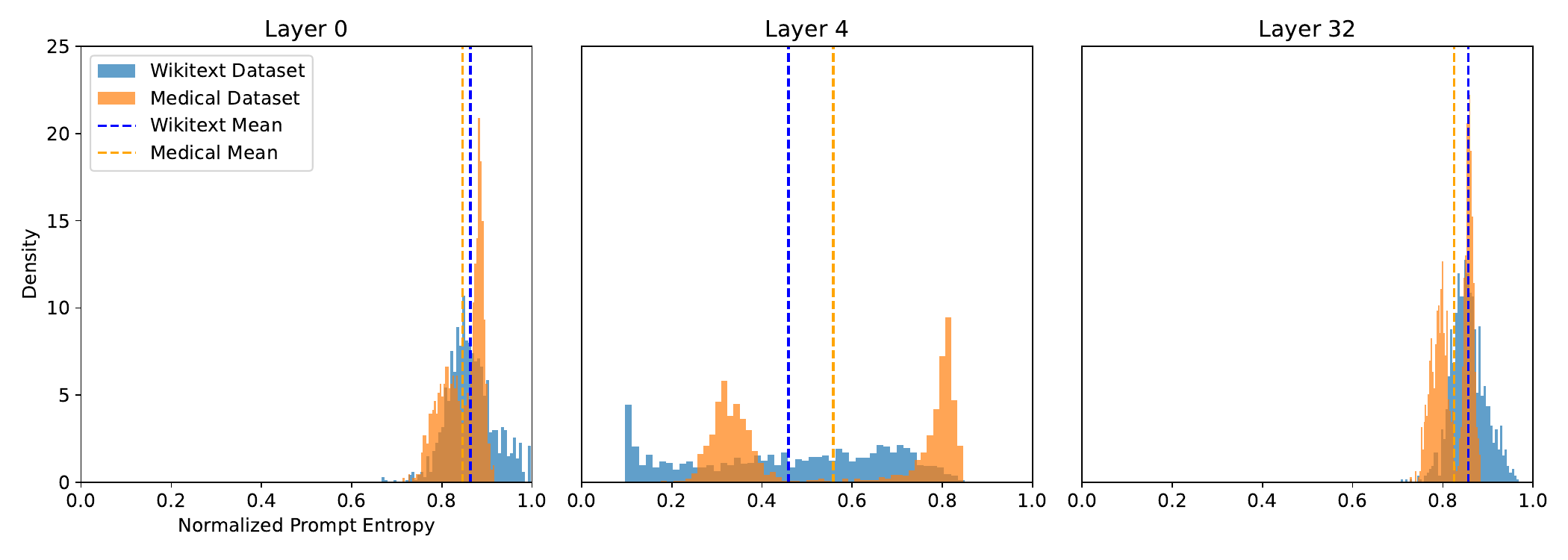}        
                \caption{Llama3 8B}
        \end{subfigure}
\caption{\textbf{Bimodal distribution of prompt entropies observed in intermediate layers.} The distributions of prompt entropies for WikiText and ai-medical-chatbot datasets are shown for Pythia, Mamba, and Llama3 models. The middle column highlights the layer with the highest Dip Test score \citep{hartigan1985dip}, which measures the degree of multimodality in the entropy distribution.}
\label{fig:all-models-bimodal}   
\end{figure}

\begin{figure}[h]
  \centering
  \includegraphics[width=\linewidth]{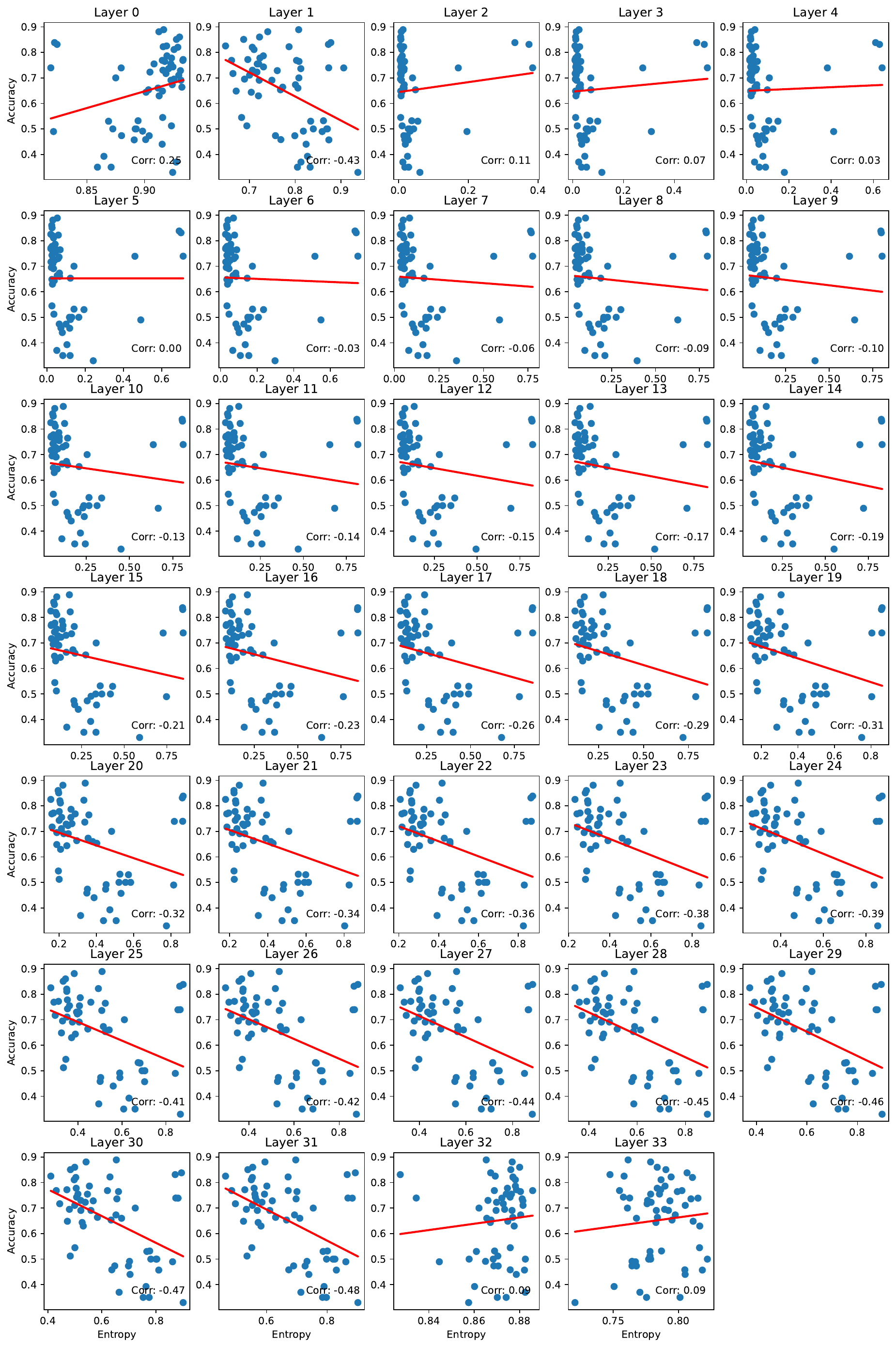}
  \caption{Entropy vs Accuracy of LLama3-8B on MMLU tasks. Each point represents a task in MMLU}
  \label{fig:with_logit22}
\end{figure}

\begin{figure}[h]
  \centering
  \includegraphics[width=0.58\linewidth]{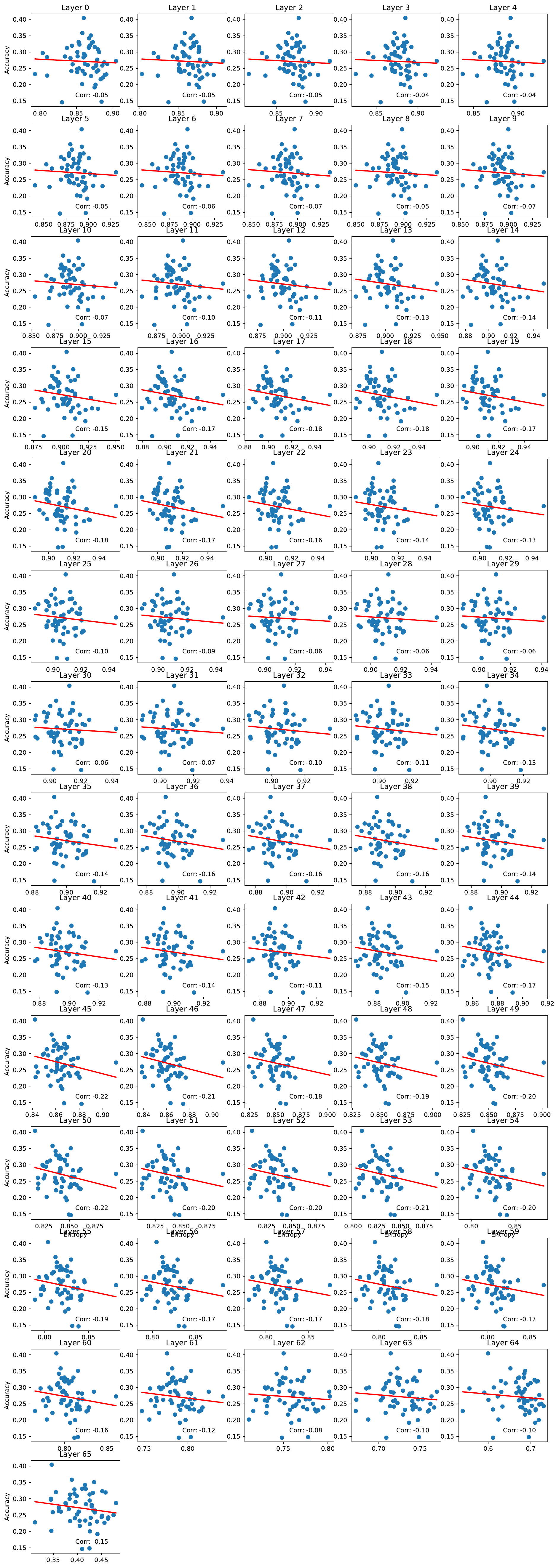}
  \caption{Entropy vs Accuracy of Mamba2-8B on MMLU tasks}
  \label{fig:with_logit11}
\end{figure}

\section{Investigation into Bimodal Distribution of Entropies}
\label{appendix:bimodal-investigation}
To determine the underlying cause of this bimodal distribution of prompt entropies, we conducted several experiments to see if specific properties of the dataset could explain this phenomenon. Our goal was to understand whether the bimodality was related to characteristics such as prompt length, semantic complexity, or overlap with training data.

\paragraph{Effect of Prompt Length}

Initially, we hypothesized that the bimodality might be caused by variations in prompt length. If one mode corresponded to shorter prompts and the other to longer prompts, it could indicate different processing strategies. However, since the entropy values were normalized and theoretically invariant to length, this was unlikely. Upon further analysis, we confirmed that prompt length did not significantly correlate with the observed bimodality.

\paragraph{Manual Examination of Prompts}

We then manually examined prompts from each mode of the distribution to identify any distinguishing features, such as difficulty or specific types of medical terminology. Despite this effort, we found no significant differences between the prompts in either mode. Both modes contained a similar range of medical complexity and varied use of terminology, suggesting that the model's entropy was not merely a reflection of the difficulty or specificity of the input.

\paragraph{Training Set Overlap}

Next, we investigated whether the low entropy mode might be associated with prompts that were very similar to samples seen during training. Given that both the ai-medical-chatbot dataset and PILE \citep{gao2020pile} (which Mamba, Pythia, and possibly Llama3 were trained on) contained medical articles from PubMed, we hypothesized that overlap with training data could lead to more confident, lower-entropy representations. To test this, we implemented a BM25 index \citep{bm25s} to quickly search for identical or highly similar articles between the two datasets.

While we did find identical articles between the ai-medical-chatbot dataset and PILE, these articles were evenly distributed across both modes of the bimodal entropy distribution. This suggests that the presence of training set overlap does not explain the bimodal behavior, and the underlying cause remains an open question.

\section{Architectural Details}
\label{appendix:architectures}

In this section, we elaborate on the specific architectures of Transformers and State Space Models (SSMs). We outline the mathematical foundations, including the weight matrices, attention mechanisms for Transformers, and the state transition matrices for SSMs. Detailed equations and parameter configurations are provided to facilitate replication and deeper understanding.

\subsection{Transformer}
The Transformer architecture \citep{vaswani2017attention} utilizes self-attention mechanisms. Given an input $\mathbf{x}$, the key ($\mathbf{K}$), query ($\mathbf{Q}$), and value ($\mathbf{V}$) matrices are computed as:

\begin{equation}
    \mathbf{Q} = \mathbf{x}\mathbf{W}_Q, \quad \mathbf{K} = \mathbf{x}\mathbf{W}_K, \quad \mathbf{V} = \mathbf{x}\mathbf{W}_V,
\end{equation}

where $\mathbf{W}_Q, \mathbf{W}_K \in \mathbb{R}^{d \times d_k}$ and $\mathbf{W}_V \in \mathbb{R}^{d \times d_v}$ are learned weights.

The attention weights are calculated using:

\begin{equation}
    \mathbf{A} = \operatorname{softmax}\left(\frac{\mathbf{Q}\mathbf{K}^\top}{\sqrt{d_k}} + \mathbf{M}\right),
\end{equation}

where $\mathbf{M}$ is a mask to enforce causality in autoregressive tasks.

The output is then:

\begin{equation}
    \mathbf{y} = \mathbf{A}\mathbf{V}.
\end{equation}

\section{Discussion on Prompt Entropy}
\label{sect:appendix-prompt-entropy}

The first measure of token embedding diversity we call prompt entropy. This entropy is measured on the intermediate tokens and captures how diverse the token representations are.

We follow the work of \cite{wei2024large} and use $\alpha$-order matrix-based entropy \cite{giraldo2014measures, skean2023dime, skean2024frossl}, which serves as a tractable surrogate for traditional Rényi’s $\alpha$-order entropy \cite{renyi1961measures}. The quantity is calculated using a similarity kernel $\kappa$ on a batch of samples drawn from a distribution, without making explicit assumptions on what the true distribution is. The choice of kernel $\kappa$ is flexible and can be any infinitely divisible kernel such as the Gaussian kernel, linear kernel, or Laplacian kernel, among others. For this work, we restrict ourselves to the linear kernel $\kappa(a, b) = a b^T$. This choice is motivated by the linear representation hypothesis \cite{parklinear2024} which finds that large language model representations encode high-level concepts such as truth \cite{burns2022dl}, honesty \cite{mallen2024eliciting}, and part-of-speech \cite{mamou2020emergence} in linearly separable manifolds.

 The equation for matrix-based entropy was previously defined in Eq. \ref{eq:matrix-based-entropy}. One way to interpret Eq. \ref{eq:matrix-based-entropy} is as the $\alpha$-order Rényi entropy of the Gram matrix eigenvalues\footnote{The non-zero eigenvalues of the Gram matrix $Z Z^T$ are equivalent to those of the covariance matrix $Z^T Z$. Using the covariance matrix instead of the Gram matrix in Eq. \ref{eq:matrix-based-entropy} makes no difference and is more computationally efficient if $D < N$.}. Notice how each eigenvalue is divided by $\textrm{tr}(\mathbf{K}_{\mathbf{Z}})$ before being raised to the $\alpha$ power. This is so that the eigenvalues of $\mathbf{K}_{\mathbf{Z}}$ sum to one (because  $\textrm{tr}(\cdot) = \sum_{i=1}^n \lambda_i(\cdot)$), which is a necessary condition to treat the eigenvalues as a probability distribution. Futhermore, each eigenvalue of $\mathbf{K}_{\mathbf{Z}}$ signifies the variance of samples in a particular principal component direction~\cite{scholkopf2018learning}. If entropy is low, then the eigenvalues form a heavy-tail distribution which implies that a few components dominate the variance of samples in $Z$. On the other hand, at maximum entropy, the eigenvalues form a uniform distribution and samples are spread equally in all directions. Matrix-based entropy is reminiscent of the LogDet entropy which uses the determinant of $\mathbf{K}_{\mathbf{Z}}$ to capture how much "volume" a dataset occupies~\cite{shwartz2023information, zhouyin2021understanding}. The LogDet entropy is given by $S_{\textrm{LogDet}}(Z) = \log \det (\mathbf{K}_{\mathbf{Z}}) - \log 2$. One can use Jensen's inequality to show that the LogDet entropy is a lower bound of Eq \ref{eq:matrix-based-entropy} when $\lim_{\alpha \rightarrow 1}$ (Appendix J.4 of~\cite{shwartz2023information}).
 
 Depending on the choice of $\alpha$, several special cases of matrix-based entropy can be recovered. In particular, when $\lim_{\alpha \rightarrow 1}$ it equals Shannon entropy (also referred to as von Neumann entropy in quantum information theory \cite{bach2022information, boes2019neumann}), and when $\alpha=2$ it equals collision entropy. Interestingly, the case of $\alpha=2$ can be calculated without explicit eigendecomposition \cite{skean2024frossl}. We show in the Appendix Figure \ref{fig:power_law_entropy} how varying values of $\alpha$ affect the matrix-based entropy of Gram matrices with eigenvalues distributed with a $\beta$-power law such that $\lambda_i = i^{-\beta}$. It is shown that for larger values of $\alpha$, smaller eigenvalues contribute more to the entropy.




\subsection{State Space Models}
\label{sec:ssm}

SSMs \citep{mamba} model sequences using recurrent dynamics. The hidden state $\mathbf{h}_t$ and output $\mathbf{y}_t$ at time $t$ are updated as:

\begin{align}
    \mathbf{h}_t &= \mathbf{A}\mathbf{h}_{t-1} + \mathbf{B}\mathbf{x}_t, \\
    \mathbf{y}_t &= \mathbf{C}\mathbf{h}_t + \mathbf{D}\mathbf{x}_t,
\end{align}

where $\mathbf{A} \in \mathbb{R}^{n \times n}$, $\mathbf{B} \in \mathbb{R}^{n \times d}$, $\mathbf{C} \in \mathbb{R}^{d \times n}$, and $\mathbf{D} \in \mathbb{R}^{d \times d}$ are learned parameters.




\section{Behavior of Matrix-based Entropy for different choices of $\alpha$}
\label{appendix:entropy}
 
 Depending on the choice of $\alpha$, several special cases of matrix-based entropy can be recovered. In particular, when $\lim_{\alpha \rightarrow 1}$ it equals Shannon entropy (also referred to as von Neumann entropy in quantum information theory \cite{bach2022information, boes2019neumann}), and when $\alpha=2$ it equals collision entropy. Interestingly, the case of $\alpha=2$ can be calculated without explicit eigendecomposition \cite{skean2024frossl}. We show in the Appendix Figure \ref{fig:power_law_entropy} how varying values of $\alpha$ affects the matrix-based entropy of Gram matrices with eigenvalues distributed with a $\beta$-power law such that $\lambda_i = i^{-\beta}$. It is shown that for larger values of $\alpha$, smaller eigenvalues contribute more to the entropy.

\begin{figure}[!b]
  \begin{center}
      \includegraphics[width=0.4\linewidth]{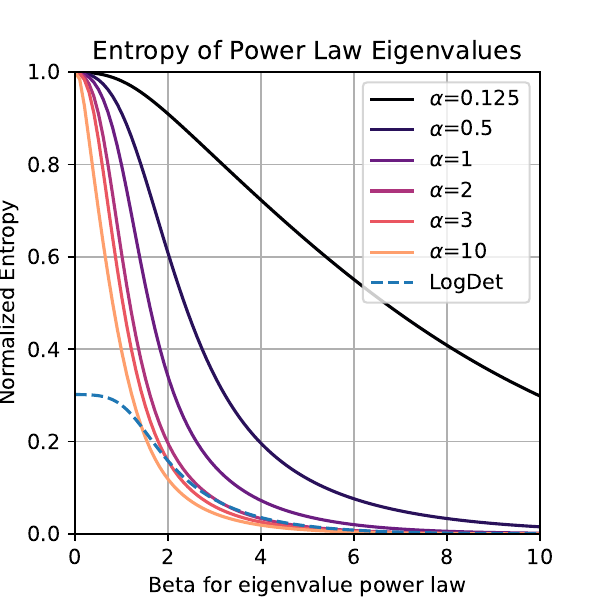}
  \end{center}
  \caption{The behavior of Eq. \ref{eq:matrix-based-entropy} for varying values of $\alpha$ on Gram matrices with eigenvalues distributed with a $\beta$-power law such that $\lambda_i = i^{-\beta}$.}
  \label{fig:power_law_entropy}
\end{figure}

\section{Dataset Details}
\label{appendix:dataset-details}
\subsection{Wikitext Dataset}
We used the wikitext dataset \cite{merity2016pointer} for the majority of our experiments in Section \ref{sect:metrics-experiments}. This was downloaded from \textbf{Salesforce/wikitext} on huggingface. The dataset consists of 100 million tokens scraped from the Featured articles on wikipedia. We filtered out prompts which were less than 30 tokens or were wikipedia section headings.

\subsection{AI-Medical-Chatbot Dataset}
We also used the medical instruction dataset called ai-medical-chatbot \cite{ruslanmv2024} which downloaded from \textbf{ruslanmv/ai-medical-dataset} on HuggingFace. An example from this dataset is:

\begin{lstlisting}
    You are an AI Medical Assistant Chatbot, trained to answer medical questions. Below is an instruction that describes a task, paired with an response context. Write a response that appropriately completes the request.
    
    ### Instruction:
    What is the resurgent sodium current in mouse cerebellar Purkinje neurons?

    ### Context:
    FGF14 modulates resurgent sodium current in mouse cerebellar Purkinje neurons.
\end{lstlisting}

\section{Prompt Augmentations}
\label{appendix:prompt-augmentation}
For the augmentation-invariance metrics such as infoNCE, LiDAR, and DiME, we use the NLPAug library \cite{ma2019nlpaug} to augment our prompts. We use three types of augmentations.

\begin{itemize}
 \item The SplitAug augmentation randomly splits words into two parts by adding a space. 
 \item The RandomCharAug augmentation randomly inserts, substitutes, swaps, or deletes characters.
 \item The Keyboard augmentation randomly substitutes characters with other characters that are at a distance of one as measured on a QWERTY keyboard. For instance, the character "k" may be replaced with "i", "l", "m", or "j".
\end{itemize}

We use the pseudocode below to do our augmentations using three types of augmentations, using the default library settings for each type. When computing augmentation-invariance metrics like infoNCE or DiME, we use the two augmented prompts rather than using one augmented prompt alongside the original prompt. Note that these augmentations may change the token length $T$ of a prompt.

\begin{verbatim}
    
    aug = naf.Sequential([
        naw.SplitAug(p=0.3),
        nac.RandomCharAug(p=0.3),
        nac.KeyboardAug(p=0.3),
    ])
    (aug_A, aug_B) = aug.augment(prompt, num_augmentations=2)

    prompt -> "The quick brown fox jumps over the lazy dog."

    aug_A ->  "The quDUk b rown fox wEmps o ver the l azy dog."
    aug_B ->  "The qTuXi bro wn fox uVm)s ob3r the la_k dog."
\end{verbatim}

\section{Extreme Prompts}
\label{appendix:extreme-prompts}

\subsection{Increasing Repetition}
We take regular prompts from the wikitext dataset, tokenize them, and then for each token we randomly replace it with probability $p$. We draw replacements tokens by sampling a random token from within the prompt. We show examples below for varying levels of $p$.

\begin{itemize}
    \item ($p = 0$) \hspace{3pt} Mint records indicate the first gold dollars were produced on May 7...
    \item ($p = 0.1$) Mint records indicate the first gold dollars were Mint Mint May 7...
    \item ($p = 0.5$) Mint records Mint Mint Mint gold dollars were Mint Mint Mint 7...
    \item ($p = 1.0$) Mint Mint Mint Mint Mint Mint Mint Mint Mint Mint Mint Mint Mint...
\end{itemize}

\subsection{Increasing Randomness}
We take regular prompts from the wikitext dataset, tokenize them, and then for each token we randomly replace it with probability $p$. We draw replacements uniformly from the tokenizer distribution. We show examples below for varying levels of $p$. Unlike the character-level random noise added to prompts in Section {with random noise discussed in Appendix \ref{appendix:prompt-augmentation} which might change the number of tokens $T$ of the prompt, the token-level random noise used here does not do so.

\begin{itemize}
    \item ($p = 0$) \hspace{3pt} Mint records indicate the first gold dollars were produced on May 7...
    \item ($p = 0.1$) Mint records indicate salivary first gold dollars were produced on May NaCl...
    \item ($p = 0.5$) Mint records Dallas actively first dollars persufors on Mayder129 18...
    \item ($p = 1.0$) arf emulsion minorensteinorianmega\_TOStack potsRecip Installifykeeping...
\end{itemize}

\subsection{Random Prompts with Certain Length}

To make a random prompt of a specific length $T$, we sample $T$ tokens uniformly from the Pythia tokenizer distribution. Such a prompt may look like the following for $T=16$: "Proposition Sequencespecific Exp fibers brows Club overviewNos toss Thinking traderMulti indoorlis".

We show how random prompt representations evolve over Pythia training checkpoints in Figure \ref{fig:training_increasing_repetition}. The random prompts we use are of length 512 tokens. It is readily observed that the prompt entropy is flat across layers in the beginning of training. As training progresses, the model compresses more and more near the final layers.

\begin{figure}[h]
  \centering
  \includegraphics[width=0.5\linewidth]{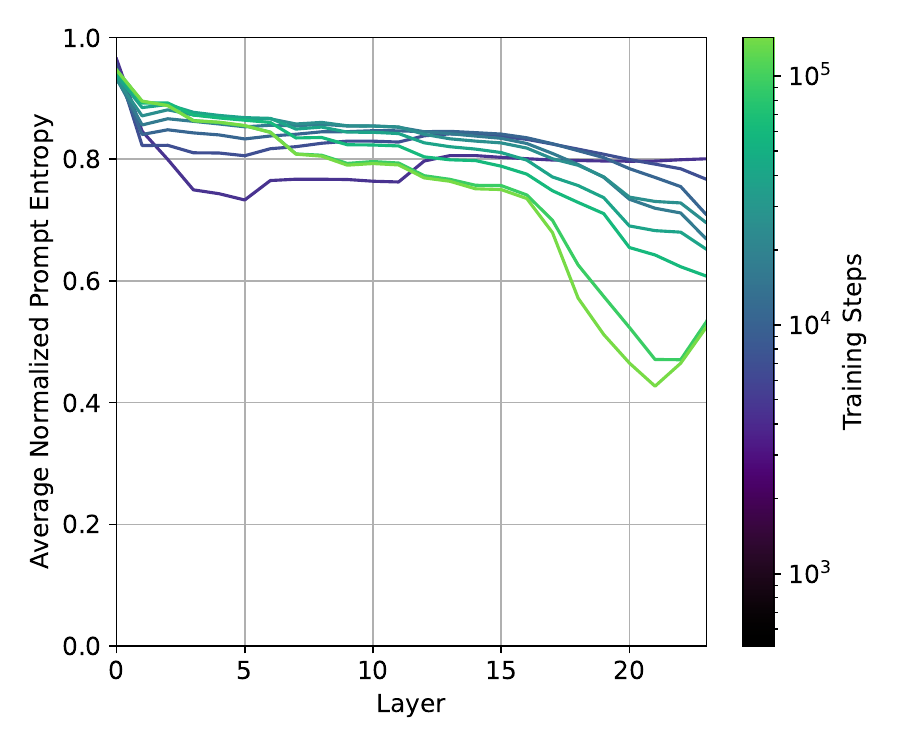}
  \caption{Behavior of random prompt representations as model is training}
\label{fig:training_increasing_repetition}
\end{figure}

\end{document}